\pdfoutput=1

\documentclass[11pt]{article}

\usepackage[final]{acl}

\usepackage{times}
\usepackage{latexsym}
\usepackage{makecell}
\usepackage{multirow}
\usepackage{multicol}
\usepackage{graphicx}
\usepackage{subfigure}
\usepackage{amssymb}
\usepackage{footnote}

\usepackage{floatrow}
\newfloatcommand{capbtabbox}{table}[][\FBwidth]
\usepackage[T1]{fontenc}

\usepackage[utf8]{inputenc}

\usepackage{microtype}

\usepackage{inconsolata}

\title{AISPACE at SemEval-2024 task 8: A Class-balanced Soft-voting System for Detecting Multi-generator Machine-generated Text}

\author{Renhua Gu, Xiangfeng Meng \\
        Samsung R\&D Institute China-Beijing \\
        \{renhua.gu, xf.meng\}@samsung.com}

\begin{document}
\maketitle
\begin{abstract}
SemEval-2024 Task 8 provides a challenge to detect human-written and machine-generated text. There are 3 subtasks for different detection scenarios. This paper proposes a system that mainly deals with Subtask B. It aims to detect if given full text is written by human or is generated by a specific Large Language Model (LLM), which is actually a multi-class text classification task. Our team AISPACE conducted a systematic study of fine-tuning transformer-based models, including encoder-only, decoder-only and encoder-decoder models. We compared their performance on this task and identified that encoder-only models performed exceptionally well. We also applied a weighted Cross Entropy loss function to address the issue of data imbalance of different class samples. Additionally, we employed soft-voting strategy over multi-models ensemble to enhance the reliability of our predictions. Our system ranked top 1 in Subtask B, which sets a state-of-the-art benchmark for this new challenge.
\end{abstract}

\section{Introduction}
Large Language Models (LLMs) have emerged as a foundational element for artificial intelligence (AI) applications. Their text generation capabilities are impressive and have almost reached the human-level performance. However, their widespread use also poses risks. The use of LLM-generated text can lead to the spread of inaccurate information, academic dishonesty, and privacy breaches. Additionally, the machine-generated text may become trapped in a loop during the process of LLMs' own development, gradually replacing human-written training data and reducing the quality and diversity of subsequent models\citep{wu2023survey}. To prevent the misuse of LLMs and improve the iterative refinement of AI tools, it is crucial to distinguish between machine-generated and human-written text. SemEval-2024 Task 8: Multigenerator, Multidomain and MultiLingual Black-Box Machine-Generated Text Detection\citep{semeval2024task8} introduces the task of detecting machine-generated text across various generators, domains and languages. Our system focuses on Subtask B, which is a multi-class classification task. It involves detecting text from multi-generators over multi-domains in English only. Given a text, the system tells whether the text is written by a human or generated by a particular LLM. It emphasizes not only the accuracy of detecting the in-domain texts, but also the generalization to identify other out-of-domain text sources.

Current research on LLMs text detection primarily focuses on ChatGPT or a specific model in a limited domain. \citealp{gao2023comparing} compares scientific writing between humans and ChatGPT exclusively. \citealp{wang2023implementing} detects AI-generated news by ChatGPT. However, there are many other emerging LLMs that generate various domain texts that needed to be distinguished from those written by humans. \citealp{wang2023m4} presents a large-scale corpus generated by popular LLMs, including ChatGPT, Cohere, Davinci, Bloomz, and Dolly, across various domains such as Wikihow, Wikipedia, Arxiv, PeerRead, and Reddit. To address this complex scenario, we fine-tune transformer-based encoder-only, decoder-only, and encoder-decoder models.We then ensemble the model that performs best for a specific class and mitigate sample imbalance using a weighted loss function. 

Our contributions can be summarized as follows:
\begin{itemize}
    \item[1)] 
    We conducted a systematic research of fine-tuning language models to detect multi-class machine-generated text.
    \item[2)]
    We developed class-balanced loss function and soft voting model ensemble to keep model robustness and generalization.
    \item[3)] 
    Our system formulated a SOTA benchmark on the task.
\end{itemize}

\section{Related Work}
Researchers have explored automatic detection methods for distinguishing machine-generated text from human-written text. These methods can be categorized into two distinct groups, i.e., metric-based methods and model-based methods \citep{he2023mgtbench}. 

\subsection{Metric-based Methods}
Metric-based methods utilize metrics such as log-likelihood, word rank, and predicted distribution entropy. For example, GLTR \citep{gehrmann2019gltr} is developed as a visualization tool to facilitate the labeling process of whether a text is machine-generated. DetectGPT \citep{mitchell2023detectgpt} define a new curvature-based criterion for distinguish machine-generated text under the assumption that text sampled from an LLM tends to occupy negative curvature regions of the model’s log probability function. GPT-who \citep{venkatraman2023gpt} is a system that computes interpretable Uniform Information Density (UID) features based on the statistical distribution of a given text. Additionally, it autonomously learns the threshold between different authors using Logistic Regression. 

\subsection{Model-based Methods}
On the other hand, model-based methods involve training classification models using both machine-generated text and human-written text. COCO \citep{liu2022coco} incorporates coherence information into text representations through the use of a graph-based encoding method. This approach is combined with a contrastive learning framework, and an enhanced contrastive loss function is proposed to mitigate potential performance degradation resulting from simple samples.

\section{System Overview}
Based on the analysis of the task situation, we have carried out preliminary studies of several methods and integrated pre-trained language models fine-tuning, class-balanced weight loss function, and soft-voting model ensemble into our system.

\subsection{Data Process}
Subtask B shares same generators, same domains and same language with subtask A. The statistical analysis reveals that subtask B lacks training data from PeerRead Source while subtask A can provide necessary data to fill the gap. To strengthen data source for training, we merged A and B train data into a unified dataset, removing any duplicated items and those present in the dev set. We then relabeled all the texts based on task B labels. The resulting training data consists of 127,755 items. For each class, the number of sample is shown in Table \ref{tab:Table 1}. However, it is important to note that the training data does not include any PeerRead texts generated by BLOOMZ, unlike the dev data. We can still assess the model's generalization ability using this data.
\begin{table}[ht!]
    \centering
    \setlength{\tabcolsep}{1.3mm}{
    \begin{tabular}{c|c|c|c|c|c}
       \hline 
       $C_{0}$&$C_{1}$&$C_{2}$&$C_{3}$&$C_{4}$&$C_{5}$\\
       \hline
        63,351&13,839&13,178&13,843&9,998&13,546\\
        \hline
    \end{tabular}
    }
    \caption{Sample Number of each class. $C_{0}$: human, $C_{1}$: ChatGPT, $C_{2}$: Cohere, $C_{3}$: Davinci, $C_{4}$: BLOOMZ, $C_{5}$: Dolly}
    \label{tab:Table 1}
\end{table}

\begin{figure}[ht!]
    \centering
    \subfigure[train]{
    \includegraphics[scale=0.17]{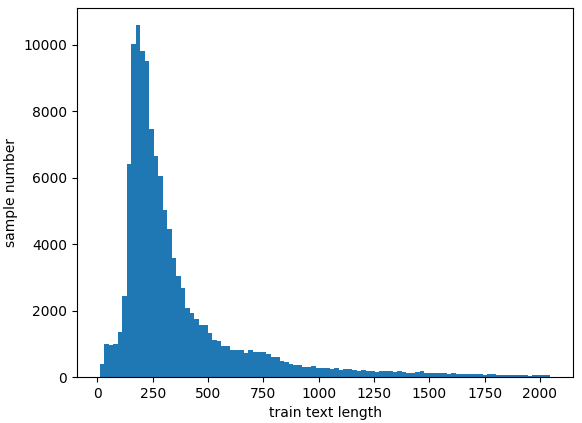}
    }
    \subfigure[dev]{
    \includegraphics[scale=0.17]{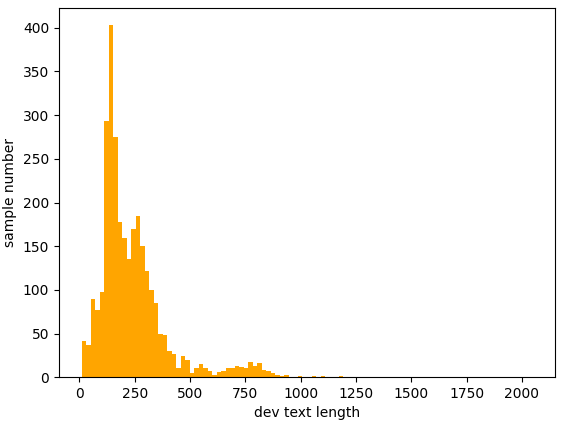}
    }
    \caption{Token length of data}
    \label{fig:Figure 1}
\end{figure}

Furthermore, after the merging of data, we analysed the token length of the dataset. As illustrated in Figure \ref{fig:Figure 1}, the majority of the token length in the training text falls within the range of 0-1000, whereas the length of the development text is mostly between 0-500. So our system tested input size of 512 and 1024 tokens in Longformer model.

\subsection{Fine-tuning Transformer-based Models}
Fine-tuning pre-trained models is typically effective approach for downstream tasks\citep{kalyan2021ammus}. Our system utilize a series of Transformer-based models, including encoder-based, decoder-based and encoder-decoder models, to develop a multi-class classifier through fine-tuning. One purpose is to determine which architecture is better suited in such tasks. Another purpose is to construct more bases that excel in different generators, which will benefit the overall ensemble results.

\subsubsection{Encoder-only}
\paragraph{Roberta} \citep{liu2019roberta} is based on the architecture of BERT \citep{devlin2018bert} and incorporates several modifications and enhancements to improve its performance. One key difference with Roberta-large is its training process, which involves training on more data for a longer period of time compared to BERT. Additionally, Roberta-large uses dynamic masking during training, where the masking pattern changes from epoch to epoch, leading to better generalization. Roberta-large has demonstrated state-of-the-art performance on various NLP benchmarks and tasks, showcasing its effectiveness in understanding and processing human language. It has been widely adopted in academic research and industry applications due to its impressive results. The model was adopted as the baseline model to explore the effectiveness of our proposed methods.

\paragraph{Deberta} \citep{he2020deberta} improved its performance from BERT by disentangled attention mechanism, which allows the model to focus on different aspects of the input independently, enabling better understanding of long-range dependencies and capturing complex linguistic structures more effectively. In addition, Deberta incorporates a novel masking scheme and dynamic upsampling during training, leading to improved model learning and generalization capability.

\paragraph{Longformer} Traditional NLP models like BERT are designed to handle sequences of up to 512 tokens, limiting their applicability to longer documents such as scientific papers, legal contracts, or lengthy news articles. Longformer \citep{beltagy2020longformer} includes a combination of global attention and sparse attention patterns. Global attention allows the model to capture relationships between distant tokens in the input sequence, while sparse attention reduces the computational complexity of processing long sequences. This balance enables Longformer to efficiently handle lengthy documents while maintaining strong performance. According to the analysis of token length, we fine-tuned 2 versions of this model with the input sizes as 512 and 1024 to asses the impact of input size. 

\subsubsection{Decoder-only}
\paragraph{XLNet} \citep{yang2019xlnet} builds upon the Transformer-XL \citep{dai2019transformer} architecture, which includes techniques for handling long-range dependencies in sequences more effectively than standard transformer architectures. This architecture enhances XLNet's ability to capture complex relationships within text data. XLNet introduces permutation language modeling, which enables the model to capture bidirectional context without relying on the autoregressive property found in traditional models like GPT-2. This approach allows XLNet to consider all permutations of the input sequence during training, leading to a more comprehensive understanding of the contextual information.

\subsubsection{Encoder-decoder}
\paragraph{T5} \citep{raffel2020exploring} belongs to the family of transformers using "text-to-text" framework, which means that it can perform a wide range of NLP tasks by converting both the input and output into text strings. This flexibility allows T5 to handle various tasks such as translation, summarization, question-answering, and more, all within a unified framework. Besides, T5 is pre-trained using a large-scale dataset and fine-tuned for specific NLP tasks, making it a highly adaptable and efficient model for a wide range of applications.

\subsection{Class Balanced Weighted Loss}
As shown in Table \ref{tab:Table 1}, each class has a different number of samples. The number of human-written samples number is even 5-6 times greater than others. To address the sample imbalance of different classes, we employed a weighted loss function during training to balance the contribution of each class sample to the loss.

For multi-class classification, the commonly used loss function is ordinary cross-entropy (CE). However when there is an imbalance-sample problem, the class-balanced weighted cross-entropy (WCE) will significantly improve the performance\citep{cui2019class}.

The weight of each class is calculated as the inverse number of samples. Denote that the number of classes is $C$, total number of all samples is $N_{total}$ the number of text samples in $Class_{i}$ is $N_{i}$, the weight factor of each class is calculated as: 

\begin{small}
\[
    \left\{
    w_{0}, w_{1}, ..., w_{C}
    \right\} = 
    \left\{
    \frac{N_{total}}{N_{0}*C}, \frac{N_{total}}{N_{1}*C}, ..., \frac{N_{total}}{N_{C}*C}
    \right\} 
\]
\end{small}

\subsection{Soft Voting}
To enhance robustness and stability across generators and domains, we employ an ensemble approach by using the soft voting method with multiple base models. 

Firstly, we obtain the confusion matrix of each base classifier. Secondly, we select the model that outperforms in a specific class. 
Finally, we integrate all the soft-max probability distribution matrix of all outperformed models to obtain the average probability distribution, and make the final decision based on it.
\begin{figure}[ht!]
    \centering
    \includegraphics{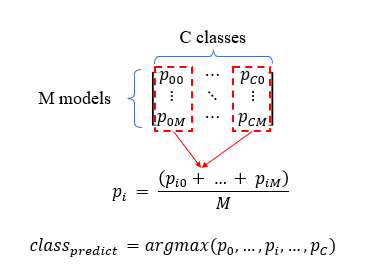}
    \caption{Soft voting over M models}
    \label{fig:Figure 2}
\end{figure}
The probability of the text belonging to $Class_{i}$ as predicted by  $Model_{j}$ is denoted by $p_{ij}$. The final probability of $Class_{i}$ is calculated as the average probabilities from M models, as illustrated in Figure \ref{fig:Figure 2}. The prediction result of the model ensemble is determined by identifying the highest probability.

\section{Experimental setup}

\subsection{Dataset and Evaluation Metrics}
For the baseline experiments on Roberta-large, we utilized the official subtaskB dataset and the merged data for separate training to determine the effectiveness of the merged data. For fine-tuning other models, we only used the merged data, which has been proven effective in the baseline experiment. The original subtaskB training data consists of 71027 items, while merging data results in a total of 127,755 items. We divided the data into train and val sets in an 8:2 ratio, and the original subtaskB dev set was kept as the dev set, which contains 3,000 items. No additional data was used for either training or evaluation. The official evaluation metric for the SubtaskB is accuracy. The experiments result in this paper is based on accuracy of dev set.

\subsection{Base models}
\paragraph{Roberta-large} serves as the baseline in our system to verify the contribution of our proposed methods. We explored different datasets, loss functions, learning rates and epochs on Roberta-large to identify which are suitable for this task. Once identified, we applied them to fine-tune other models further.

There are groups of experiments on Roberta-large, settings are as follows:
\begin{itemize}
    \item Dataset contribution: The model was fine-tuned with the  original data and merged data for 3 epochs using a learning rate of 3e-5.
    \item Loss function contribution: The model was fine-tuned with merged data for 3 epochs with learning rate of 3e-5, applying ordinary CE and weighted CE loss in the process of training.
    \item Epoch contribution: The model was fine-tuned with the merged data for 3 and 5 epochs with a learning rate of 2e-5.
    \item Learning rate contribution: The model was fine-tuned with merged data for 5 epochs with learning rates of 1e-5, 1.5e-5, 2e-5 and 3e-5.
\end{itemize}
\paragraph{Deberta-large and XLNet-large} applied the merged data and weighted CE as they have been shown to be effective in previous experiments. We explored various learning rates (including 1e-5, 2e-5, and 3e-5) and epochs (including 2epochs, 3 epochs, and 5 epochs), and selected the best setting as a learning rate of 1e-5 and 3 epochs as the performance comparison to other models.

\paragraph{Longformer} is good at handling long documents, breaking the limits of 512 tokens of Bert-family models. Since we have part of long documents whose tokens numbers are greater than 512 tokens, so we tried 1024 tokens input as well as 512 tokens. Further more, to keep the pre-trained ability on semantic understanding, we fixed the top 18 layers and only fine-tuned the remained ones. Then we fine-tuned it with merged data for 5 epochs with a learning rate of 3e-5.

\begin{table*}
\begin{floatrow}
\capbtabbox{
    \begin{tabular}{l|c|c}
        \hline
        Method&Epochs/LR&Accuracy\\
        \hline
        baseline&3/3e-5&0.7390\\
        + \begin{small} merged data\end{small}&3/3e-5&0.9050\\
        + \begin{small} merged data + WCE\end{small}&3/3e-5&0.9150\\
        + \begin{small} merged data + WCE\end{small}&5/3e-5&0.9733\\
        + \begin{small} merged data + WCE\end{small}&5/2e-5&\textbf{0.9800}\\
        + \begin{small} merged data + WCE\end{small}&5/1.5e-5&0.9626\\
        + \begin{small} merged data + WCE\end{small}&5/1e-5&0.9433\\
        \hline
    \end{tabular}
}{
 \caption{The performance Comparison of multiple methods on Roberta-large}
 \label{tab: Table 2}
}
\capbtabbox{
    \begin{tabular}{l|l|c}
        \hline
        \begin{small}
            Architecture
        \end{small}&Model & Accuracy \\
        \hline
        \multirow{4}{*}{Encoder}&Roberta-large& \textbf{0.9800}\\
         &Deberta-large&0.9730\\
         &Longformer-512&0.9643\\
         &Longformer-1024&0.9573\\
        \cline{1-2}
        Decoder&XLNet&0.9730\\
        \cline{1-2}
        \makecell[l]{Encoder\\-Decoder}&T5&0.8617\\
        \hline
    \end{tabular}
}{
 \caption{The performance comparison of different base models}
 \label{tab:Table 3}
 \small
}
\end{floatrow}
\end{table*}

\begin{table*}[ht!]
    \centering
    \begin{tabular}{l|c|c|c|c|c|c|c}
        \hline
        \multirow{2}{*}{ensemble base models}&\multicolumn{6}{c|}{excel in}&\multirow{2}{*}{accuracy}\\ 
        \cline{2-7}
        
 &$Class_{0}$&$Class_{1}$&$Class_{2}$&$Class_{3}$&$Class_{4}$&$Class_{5}$& \\
        \hline
        best single model&\multicolumn{6}{c|}{}&0.9800\\
        \hline
        Roberta-large&\checkmark&\checkmark& &\checkmark& & &\multirow{2}{*}{0.9913}\\
        Deberta-large& &\checkmark& & &\checkmark& & \\
        \hline
        Roberta-large&\checkmark&\checkmark& &\checkmark& & &\multirow{3}{*}{0.9943}\\
        Deberta-large& &\checkmark& & &\checkmark& & \\
        XLNet-large& & & & &\checkmark& & \\
        \hline
        Roberta-large&\checkmark&\checkmark& &\checkmark& & &\multirow{4}{*}{\textbf{0.9946}}\\
        Deberta-large& &\checkmark& & &\checkmark& & \\
        XLNet-large& & & & &\checkmark& & \\
        Longformer&\checkmark&\checkmark&\checkmark& & &\checkmark& \\ 
        \hline
    \end{tabular}
    \caption{The performance comparison of different base model ensemble}
    \label{tab:Table 4}
\end{table*}

\paragraph{T5} is pre-trained on a large set of corpus and has strong adaption. We fine-tuned it with our merged data over its default parameter setting for 3 epochs with learning rates at 2e-5. Referring to the appendix of T5, a prefix (M4 sentence: ) was added to each input text, then the model was trained to generate "human" or "machine".

\section{Results and Analysis}
To assess the efficacy of our proposed methods, we carried out multiple sets of experiments.

On the baseline Roberta-large, an ablation study was conducted. The performance comparison is shown in Table \ref{tab: Table 2}. The results of experiments indicate that supplementing the data source significantly improves performance. Therefore, the supervised fine-tuning is crucial in such cases. Additionally, a weighted loss function can mitigate sample imbalance issue.

Further, we fine-tuned a series of transformer-based models to select the most suitable base model. The results in Table \ref{tab:Table 3} shows that the encoder or decoder can achieve top performance while the Encoder-Decoder is poor for this task. For input size, 512 tokens exceed 1024 tokens. To include longer input has no contribution to the result. 

At last, we conducted a model ensemble by soft voting method to ensure robustness and generalization and reduce the effect of noise. The selected single base fine-tuned model is chosen based on its performance in the specific class. We tested various combinations, and the results are shown in Table \ref{tab:Table 4}. We attempted to combine various single base models, including 2, 3, and 4 types. Compared to the best single model, the ensembled model showed significant improvement, even with the least number of ensemble types. Furthermore, as the differences in ensemble models increased, the results improved even further. Additionally, if the ensemble base models perform well individually in every class, the overall result is also improved.

\section{Conclusion}
This paper presents a systematic study on detecting machine-generated text from multi-generators and multi-domains. We fine-tuned a series of transformer-based models and found that the encoder architecture is better suited for the task. We employed a weighted Cross Entropy loss function to address the sample imbalance. To improve robustness and generalization, various base models were ensembled by soft-voting method, and resulting in 99.46\% accuracy on the dev set. In the final test, our system ranked 1st. Moving forward, we plan to explore more widely used LLMs and work towards enhancing our capabilities in few-shot learning and transfer-learning for similar tasks.

\bibliography{acl_latex}

\end{document}